\documentclass{article} 
\usepackage{iclr2018_workshop,times}
\usepackage{hyperref}
\usepackage{url}

\usepackage{scrextend}
\usepackage{amsfonts}
\usepackage[utf8]{inputenc} 
\usepackage[T1]{fontenc}    
\usepackage{glossaries}
\usepackage{graphicx}
\usepackage{sidecap}
\sidecaptionvpos{figure}{c}
\usepackage{caption}
\usepackage{subcaption}
\usepackage{appendix}
\usepackage{multicol}
\usepackage{multirow}
\usepackage{booktabs}       
\usepackage{placeins}

\usepackage{tablefootnote}

\glsdisablehyper 

\def\bestcnnvaltwo{56.16} 
\def\bestrnvaltwo{72.54} 
\def\bestvggvaltwo{52.31} 
\def\besttextvaltwo{50.01} 

\def\bestrntestone{76.52} 
\def\bestrntesttwo{72.40} 
\def\bestcnntesttwo{56.00} 
\def\bestvggtesttwo{52.47} 
\def\besttexttesttwo{50.01} 
\def\besthumantesttwohuman{91.21} 
\def\bestcnntesttwohuman{56.04} 
\def\bestrntesttwohuman{72.18} 
\def\bestrnonetestone{67.74} 
\def\bestrnonetesttwo{60.35} 

\title{FigureQA: An Annotated Figure Dataset for Visual Reasoning}

\author{
Samira Ebrahimi Kahou$\mathbf{^1}$\thanks{Equal contribution.}, Vincent Michalski$\mathbf{^{2\ast}}$\thanks{Part of this work was completed while the author interned at Microsoft Research Montréal.}, Adam Atkinson$\mathbf{^1}$, \\\textbf{Ákos Kádár$\mathbf{^{3\dag}}$, Adam Trischler$\mathbf{^1}$, Yoshua Bengio$\mathbf{^3}$}\\
$^1$Microsoft Research Montréal\\
$^2$Université de Montréal, MILA\\
$^3$Tilburg University
}

\begin{document}

\newacronym{cnn}{CNN}{Convolutional Neural Network}
\newacronym{gre}{GRE}{Graduate Records Examination}
\newacronym{lstm}{LSTM}{Long Short-Term Memory}
\newacronym{mlp}{MLP}{Multi-Layer Perceptron}
\newacronym{ocr}{OCR}{optical character recognition}
\newacronym{png}{PNG}{Portable Network Graphics}
\newacronym{relu}{ReLU}{Rectified Linear Unit}
\newacronym{rn}{RN}{Relation Network}
\newacronym{rnn}{RNN}{Recurrent Neural Network}
\newacronym{sgd}{SGD}{Stochastic Gradient Descent}
\newacronym{vqa}{VQA}{Visual Question Answering}

\maketitle
\begin{abstract}
We introduce FigureQA, a visual reasoning corpus of over one million question-answer pairs grounded in over $100,000$ images.
The images are synthetic, scientific-style figures from five classes: line plots, dot-line plots, vertical and horizontal bar graphs, and pie charts.
We formulate our reasoning task by generating questions from 15 templates; questions concern various relationships between plot elements and examine characteristics like the maximum, the minimum, area-under-the-curve, smoothness, and intersection.
To resolve, such questions often require reference to multiple plot elements and synthesis of information distributed spatially throughout a figure.
To facilitate the training of machine learning systems, the corpus also includes side data that can be used to formulate auxiliary objectives. In particular,
we provide the numerical data used to generate each figure as well as bounding-box annotations for all plot elements.
We study the proposed visual reasoning task by training several models, including the recently proposed Relation Network as a strong baseline.
Preliminary results indicate that the task poses a significant machine learning challenge.
We envision FigureQA as a first step towards developing models that can intuitively recognize patterns from visual representations of data.
\end{abstract}

\section{Introduction}
Scientific figures compactly summarize valuable information. They depict patterns like trends, rates, and proportions, and enable humans to understand these concepts intuitively at a glance.
Because of these useful properties, scientific papers and other documents often supplement textual information with figures.
Machine understanding of this structured visual information could assist human analysts in extracting knowledge from the vast documentation produced by modern science. 
Besides immediate applications, machine understanding of plots is interesting from an artificial intelligence perspective, as most existing approaches simply revert to reconstructing the source data, thereby inverting the visualization pipeline. Mathematics exams, such as the \glspl{gre}, often include questions regarding relationships between plot elements of a figure. When solving these exam questions, humans do not always build a table of coordinates for all data points, but often judge by visual intuition.

Thus motivated, and inspired by recent research in \emph{\gls{vqa}}~\citep{antol2015vqa,goyal2016making} and \emph{relational reasoning}~\citep{johnson2016clevr,suhr2017corpus}, we introduce \emph{FigureQA}.
FigureQA is a corpus of over one million question-answer pairs grounded in over $100,000$ figures, devised to study aspects of comprehension and reasoning in machines.
There are five common figure types represented in the corpus, which model both continuous and categorical information: \emph{line}, \emph{dot-line}, \emph{vertical} and \emph{horizontal bar}, and \emph{pie plots}.
Questions concern \emph{one-to-all} and \emph{one-to-one} relations among plot elements, e.g. \emph{Is X the low median?}, \emph{Does X intersect Y?}. Their successful resolution requires inference over multiple plot elements.
There are 15 question types in total, which address properties like \emph{magnitude}, \emph{maximum}, \emph{minimum}, \emph{median}, \emph{area-under-the-curve}, \emph{smoothness}, and \emph{intersections}. Each question is posed such that its answer is either \emph{yes} or \emph{no}.

FigureQA is a synthetic corpus, like the related \emph{CLEVR} dataset for visual reasoning~\citep{johnson2016clevr}.
While this means that the data may not exhibit the same richness as figures ``in the wild'',  it permits greater control over the task's complexity, enables auxiliary supervision signals, and most importantly provides reliable ground-truth answers. Furthermore, by analyzing the performance on real figures of models trained on FigureQA it will be possible to extend the corpus to address limitations not considered during generation.
The FigureQA corpus can be extended iteratively, each time raising the task complexity, as model performance increases. This is reminiscent of \emph{curriculum learning} \citep{bengio2009curriculum} allowing iterative pretraining on increasingly challenging versions of the data. By releasing the data now, we want to gauge the interest in the research community and adapt future versions based on feedback, to accelerate research in this field.
Additional annotation is provided to allow researchers to define tasks other than the one we introduce in this manuscript.

The corpus is built using a two-stage generation process. First, we sample numerical data according to a carefully tuned set of constraints and heuristics designed to make sampled figures appear natural.
Next we use the \emph{Bokeh} open-source plotting library~\citep{bokeh} to plot the data in an image. This process necessarily gives us access to the quantitative data presented in the figure. We also modify the Bokeh backend to output bounding boxes for all plot elements: data points, axes, axis labels and ticks, legend tokens, etc. We provide the underlying numerical data and the set of bounding boxes as supplementary information with each figure, which may be useful in formulating auxiliary tasks, like reconstructing quantitative data given only a figure image. The bounding box targets of plot elements relevant to a question may be useful for supervising an attention mechanism, which can ignore potential distractions. Experiments in that direction are outside of the scope of this work, but we want to facilitate research of such approaches by releasing these annotations.

As part of the generation process we balance the ratio of yes and no answers for each question type and each figure. 
This makes it more difficult for models to exploit biases in answer frequencies while ignoring visual content.

We review related work in Section~\ref{sec:related}.
In Section~\ref{sec:dataset} we describe the FigureQA dataset and the visual-reasoning task in detail. Section~\ref{sec:models} describes and evaluates four neural baseline models trained on the corpus: a text-only \gls{lstm} model~\citep{hochreiter1997long} as a sanity check for biases, the same \gls{lstm} model with added \gls{cnn} image features~\citep{lecun1998gradient,fukushima1988neocognitron}, one baseline instead using pre-extracted VGG image features \citep{simonyan2014very}, and a \gls{rn}~\citep{santoro2017simple}, a strong baseline model for relational reasoning. 

The \gls{rn} achieves respective accuracies of \bestrntesttwo\% and \bestrntestone\% on the FigureQA test set with alternated color scheme (described in Section~\ref{sec:source_data_and_figures}) and the test set without swapping colors.
An ``official'' version of the corpus is publicly available as a benchmark for future research.\footnote{\url{https://datasets.maluuba.com/FigureQA}} 
We also provide our generation scripts\footnote{\url{https://github.com/Maluuba/FigureQA}}, which are easily configurable, enabling researchers to tweak parameters to produce their own variations of the data, and our baseline implementations\footnote{\url{https://github.com/vmichals/FigureQA-baseline}}.

\section{Related work}
\label{sec:related}
Machine learning tasks that pose questions about visual scenes have received great interest of late.
For example, \citet{antol2015vqa} proposed the \gls{vqa} challenge, in which a model seeks to output a correct natural-language answer $a$ to a natural-language question $q$ concerning image $I$.
An example is the question ``Who is wearing glasses?'' about an image of a man and a woman, one of whom is indeed wearing glasses.
Such questions typically require capabilities of vision, language, and common-sense knowledge to answer correctly.
Several works tackling the \gls{vqa} challenge observe
that models tend to exploit strong linguistic priors rather than learning
to understand visual content. To remedy this problem, \citet{goyal2016making} 
introduced the balanced \gls{vqa} task. This features
triples  $(I', q, a')$ to supplement each image-question-answer triple $(I, q, a)$, 
such that $I'$ is similar to $I$ but the answer given $I'$ and the 
same $q$ is $a'$ rather than $a$.

Beyond linguistic priors, another potential issue with the \gls{vqa} challenges stems from their use of real images.
Images of the real world entangle visual-linguistic reasoning with common-sense concepts,
where the latter may be too numerous to learn from \gls{vqa} corpora alone.
On the other hand, synthetic datasets for visual-linguistic reasoning may not require common sense and may permit the reasoning challenge to be studied in isolation.
CLEVR~\citep{johnson2016clevr} and \emph{NLVR}~\citep{suhr2017corpus} are two such corpora.
They present scenes of simple geometric objects along with questions concerning their arrangement. To answer such questions, machines should be capable of spatial and relational reasoning.
These tasks have instigated rapid improvement in neural models for visual understanding~\citep{santoro2017simple,perez2017learning,hu2017learning}.
FigureQA takes the synthetic approach of CLEVR and NLVR for the same purpose, to contribute to advances in figure-understanding algorithms.

The figure-understanding task has itself been studied previously.
For example, \citet{siegel2016figureseer} present a smaller dataset of figures extracted from research papers, along with a pipeline model for analyzing them.
As in FigureQA, they focus on answering linguistic questions about the underlying data. 
Their \emph{FigureSeer} corpus contains $60,000$ figure images annotated by crowdworkers with the plot-type labels.
A smaller set of 600 figures comes with richer annotations of axes, legends, and plot data, similar to the annotations we provide for all $140,000$ figures in our corpus.
The disadvantage of FigureSeer as compared with FigureQA is its limited size; the advantage is that its plots come from real data.
The questions posed in FigureSeer also entangle reasoning about figure content with several detection and recognition tasks, such as localizing axes and tick labels or matching line styles with legend entries. Among other capabilities, models require good performance in \gls{ocr}.
Accordingly, the model presented by \citet{siegel2016figureseer} comprises a pipeline of disjoint, off-the-shelf components that are not trained end-to-end.

\citet{poco2017reverse} propose the related task of recovering visual encodings from chart images. This entails detection of legends, titles, labels, etc., as well as classification of chart types and text recovery via \gls{ocr}.
Several works focus on data extraction from figures.
\citet{tsutsui2017data} use convolutional networks to detect boundaries of subfigures and extract these from compound figures;
\citet{jung2017chartsense} propose a system for processing chart images, which consists of figure-type classification followed by type-specific interactive tools for data extraction.
Also related to our work is the corpus of \citet{cliche2017scatteract}.
There, the goal is automated extraction of data from synthetically generated scatter plots. This is equivalent to the data-reconstruction auxiliary task available with FigureQA.

FigureQA is designed to focus specifically on reasoning, rather than subtasks that can be solved with high accuracy by existing tools for \gls{ocr}.
It follows the general \gls{vqa} setup, but additionally provides rich 
bounding-box annotations for each figure along with underlying numerical data.
It thus offers a setting in which existing and novel visual-linguistic models can be trained from scratch 
and may take advantage of dense supervision.
Its questions often require reference to multiple plot elements and synthesis of information distributed spatially throughout a figure.
The task formulation is aimed at achieving an ``intuitive'' figure-understanding system, that does not resort to inverting the visualization pipeline. This is in line with the recent trend in visual-textual datasets, such as those for intuitive physics and reasoning~\citep{goyal2017something,mun2016marioqa}.

The majority of recent methods developed for \gls{vqa} and related vision-language tasks, such as image captioning 
\citep{xu2015show, fang2015captions}, video-captioning~\citep{yu2016video}, phrase localization~\citep{hu2016natural}, 
and multi-modal machine translation~\citep{elliott2017imagination}, employ a neural
encoder-decoder framework. These models typically encode the visual modality with 
pretrained \glspl{cnn}, such as VGG~\citep{simonyan2014very} or ResNet
\citep{he2016deep}, and may extract additional information from images 
using pretrained object detectors~\citep{ren15fasterrcnn}. Language encoders
based on bag-of-words or \gls{lstm} approaches are typically either trained from scratch
\citep{elliott2017imagination} or make use of pretrained word embeddings~\citep{you2016image}.
Global or local image representations are typically combined with the language encodings through attention
\citep{xiong2016dynamic, yang2016stacked, lu2016hierarchical} and pooling~\citep{fukui2016multimodal} mechanisms,
then fed to a decoder that outputs a final answer in language.
In this work we evaluate a standard CNN-LSTM encoder model as well as a more recent architecture designed expressly for relational reasoning~\citep{santoro2017simple}.

\section{Dataset}
\label{sec:dataset}
FigureQA consists of common scientific-style plots accompanied by questions and answers concerning them. The corpus is synthetically generated at large scale: its training set contains $100,000$ images with 1.3 million questions; the validation and test sets each contain $20,000$ images with over $250,000$ questions.

\begin{figure}[h]
    \centering
    \caption{Sample line plot figure with question-answer pairs.}
    \label{fig:sample_line_qa}
    \begin{subfigure}[]{0.65\textwidth}
        \includegraphics[width=0.95\textwidth]{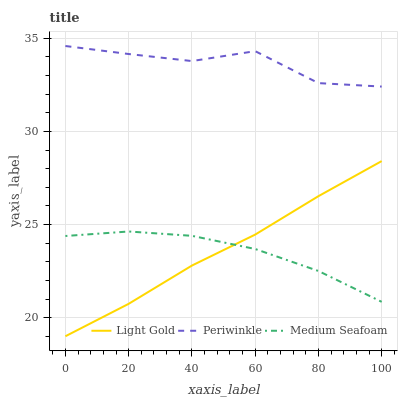}
    \end{subfigure}
    \hfill
    \begin{subfigure}[]{0.3\textwidth}
        \textbf{Q:} Does Medium Seafoam intersect Light Gold?\newline
        \textbf{A: Yes}\newline\newline
        \textbf{Q:} Is Medium Seafoam the roughest?\newline
        \textbf{A: No}\newline\newline
        \textbf{Q:} Is Light Gold less than Periwinkle?\newline
        \textbf{A: Yes}\newline\newline
        \textbf{Q:} Does Periwinkle have the maximum area under the curve?\newline
        \textbf{A: Yes}\newline\newline
        \textbf{Q:} Does Medium Seafoam have the lowest value?\newline
        \textbf{A: No}
    \end{subfigure}
\end{figure}

The corpus represents numerical data according to five figure types commonly found in analytical documents, namely, horizontal and vertical bar graphs, continuous and discontinuous line charts, and pie charts. These figures are produced with white background and the colors of plot elements (lines, bars and pie slices) are chosen from a set of $100$ colors (see Section~\ref{sec:source_data_and_figures}). Figures also contain common plot elements such as axes, gridlines, labels, and legends.
We generate question-answer pairs for each figure from its numerical source data according to predefined templates. We formulate 15 questions types, given in Table~\ref{tab:questionTypes}, that compare quantitative attributes of two plot elements or one plot element versus all others. In particular, questions examine properties like the \emph{maximum}, \emph{minimum}, \emph{median}, \emph{roughness}, and \emph{greater than}/\emph{less than} relationships. All are posed as a binary choice between \emph{yes} and \emph{no}.
In addition to the images and question-answer pairs, we provide both the source data and bounding boxes for all figure elements, and supplement questions with the names, RGB codes, and unique identifiers of the featured colors. These are for optional use in analysis or to define auxiliary training objectives.

In the following section, we describe the corpus and its generation process in depth.

\subsection{Source Data and Figures}
\label{sec:source_data_and_figures}
The many parameters we use to generate our source data and figures are summarized in Table~\ref{tab:params}.
These constrain the data-sampling process to ensure consistent, realistic plots with a high degree of variation.
Generally, we draw data values from uniform random distributions within parameter-limited ranges.
We further constrain the ``shape'' of the data using a small set of commonly observed functions (linear, quadratic, bell curve) with additive perturbations.

A figure's data points are identified visually by color; textually (on axes and legends and in questions), we identify data points by the corresponding color names.
For this purpose we chose 100 unique colors from the X11 named color set\footnote{See \url{https://cgit.freedesktop.org/xorg/app/rgb/tree/rgb.txt}.}, selecting those with a large color distance from white, the background color of the figures.

We construct FigureQA's training, validation, and test sets such that all 100 colors are observed during training, while validation and testing are performed on unseen \emph{color-plot combinations}.
This is accomplished using a methodology consistent with that of the CLEVR dataset~\citep{johnson2016clevr}, as follows.
We divide our 100 colors into two disjoint, equally-sized subsets (denoted $A$ and $B$). In the training set, we color a particular figure type by drawing from one, and only one, of these subsets (see Table~\ref{tab:params}).
When generating the validation and test sets, we draw from the \emph{opposite} subset used for coloring the figure in the training set, i.e., if subset $A$ was used for training, then subset $B$ is used for validation and testing.
We define this coloring for the validation and test sets as the  ``alternated color scheme.''\footnote{We additionally provide validation and test sets built without this scheme.}

We define the appearance of several other aspects during data generation, randomizing these as well to encourage variation.
The placement of the legend within or outside the plot area is determined by a coin flip, and we select its precise location and orientation to cause minimal obstruction by counting the occupancy of cells in a $3\times3$ grid.
Figure width is constrained to within one to two times its height, there are four font sizes available, and grid lines may be rendered or not -- all with uniform probability.

\begin{table}[h]
    \caption{Synthetic Data Parameters, with color sets used for each color scheme.}
    \label{tab:params}
    \begin{center}
        \setlength\tabcolsep{5pt} 
        \begin{tabular}{lcclcc}
            \toprule
            \multirow{2}{*}{Figure Types}    & \multirow{2}{*}{Elements}  & \multirow{2}{*}{Points}    & \multirow{2}{*}{Shapes}    & \multicolumn{2}{c}{Color Scheme}\\
                & & & & Training   & Alternated\\
            \midrule
            Vertical Bar     & 1       & 2-10         & uniform random, linear, bell-shape       & $A$ & $B$\\
            Horizontal Bar     & 1       & 2-10         & uniform random, linear, bell-shape       & $B$  & $A$ \\
            Line\tablefootnote{Lines are drawn in five styles.}     & 2-7       & 5-20         & linear, linear with noise, quadratic       & $A$  & $B$ \\
            Dot Line     & 2-7       & 5-20         & linear, linear with noise, quadratic       & $B$ & $A$ \\
            Pie     & 2-7       & 1         & N/A       & $A$   & $B$ \\
            \bottomrule
        \end{tabular}
    \end{center}
\end{table}

\subsection{Questions and Answers}
We generate questions and their answers by referring to a figure's source data and applying the templates given in Table \ref{tab:questionTypes}. One \emph{yes} and one \emph{no} question is generated for each template that applies.

Once all question-answer pairs have been generated, we filter them to ensure an equal number of yes and no answers by discarding question-answer pairs until the answers per question type are balanced. This removes bias from the dataset to prevent models from learning summary statistics of the question-answer pairs.

Note that since we provide source data for all the figures, arbitrary additional questions may be synthesized. This makes the dataset extensible for future research. 

To measure the smoothness of curves for question templates 9 and 10, we devised a roughness metric based on the sum of absolute pairwise differences of slopes, computed via finite differences. Concretely, for a curve with $n$ points defined by series $\mathbf{x}$ and $\mathbf{y}$,

\[\mathtt{Roughness}(\mathbf{x},\mathbf{y})= \displaystyle\sum\limits_{i=1}^{n-2} \left\lvert \dfrac{y_{i+2} - y_{i+1}}{x_{i+2} - x_{i+1}} - \dfrac{y_{i+1} - y_{i}}{x_{i+1} - x_{i}}\right\rvert. \]

\subsection{Plotting}
We generate figures from the synthesized source data using the open-source plotting library \emph{Bokeh}. Bokeh was selected for its ease of use and modification and its expressiveness. We modified the library's web-based rendering component to extract and associate bounding boxes for all figure elements. Figures are encoded in three channels (RGB) and saved in \gls{png} format.

\begin{table}[h]
    \caption{Question Types.}
    \label{tab:questionTypes}
    \begin{center}
        \begin{tabular}{clc}
            \toprule
            {} & Template   & Figure Types \\
            \midrule
            1 & Is $X$ the minimum?         & bar, pie \\
            2 & Is $X$ the maximum?         & bar, pie \\
            3 & Is $X$ the low median?      & bar, pie \\
            4 & Is $X$ the high median?     & bar, pie \\
            5 & Is $X$ less than $Y$?       & bar, pie \\
            6 & Is $X$ greater than $Y$?    & bar, pie \\
            7 & Does $X$ have the minimum area under the curve? & line \\
            8 & Does $X$ have the maximum area under the curve? & line \\
            9 & Is $X$ the smoothest?       & line \\
            10 & Is $X$ the roughest?        & line \\
            11 & Does $X$ have the lowest value?     & line \\
            12 & Does $X$ have the highest value?    & line \\
            13 & Is $X$ less than $Y$?\tablefootnote{\label{note:strictly} In the sense of \emph{strictly greater/less} than. This clarification is provided to judges for the human baseline.}       & line \\
            14 & Is $X$ greater than $Y$?\footref{note:strictly}    & line \\
            15 & Does $X$ intersect $Y$?     & line \\
            \bottomrule
        \end{tabular}
    \end{center}
\end{table}

\section{Models}
\label{sec:models}
To establish baseline performances on FigureQA, we implemented the four models described below. In all experiments we use training, validation, and test sets with the alternated color scheme (see Section~\ref{sec:source_data_and_figures}).
The results of an experiment with the \gls{rn} baseline trained and evaluated with different schemes is provided in Appendix~\ref{sec:early_stop}.
We train all models using the \emph{Adam} optimizer~\citep{kingma2014adam} on the standard cross-entropy loss with learning rate $0.00025$.

\paragraph{Preprocessing} We resize the longer side of each image to 256 pixels, preserving the aspect ratio; images are then padded with zeros to size $256\times256$. For data augmentation, we use the common scheme of padding images (to size $264\times264$) and then randomly cropping them back to the previous size ($256\times256)$.

\paragraph{Text-only baseline}
Our first baseline is a text-only model that uses an \gls{lstm}\footnote{The TensorFlow~\citep{abadi2016tensorflow} implementation based on the seminal work of \citet{hochreiter1997long}.} to read the question word by word. Words are represented by a learned embedding of size 32 (our vocabulary size is only 85, not counting default tokens such as those marking the start and end of a sentence). The \gls{lstm} has 256 hidden units. 
A \gls{mlp} classifier passes the last \gls{lstm} state through two hidden layers with 512 \glspl{relu}~\citep{nair2010rectified} to produce an output. The second hidden layer uses dropout at a rate of 50\%~\citep{srivastava2014dropout}. This model was trained with batch size 64.

\paragraph{CNN+LSTM}
In this model the \gls{mlp} classifier receives the concatenation of the question encoding with a learned visual representation.
The visual representation comes from a \gls{cnn} with five convolutional layers, each with 64 kernels of size $3\times3$, stride 2, zero padding of 1 on each side and batch normalization~\citep{ioffe2015batch}, followed by a fully-connected layer of size 512. All layers use the \gls{relu} activation function.
The \gls{lstm} producing the question encoding has the same architecture as in the text-only model. This baseline was trained using four parallel workers each computing gradients on batches of size 160 which are then averaged and used for updating parameters.

\paragraph{CNN+LSTM on VGG-16 features}
In our third baseline we extract features from layer \emph{pool5} of an \emph{ImageNet}-pretrained \emph{VGG-16} network \citep{simonyan2014very} using the code provided with \citet{hu2017learning}. 
The extracted features ($512$ channels of size $10\times15$) are then processed by a \gls{cnn} with four convolutional layers, all with $3\times3$ kernels, \gls{relu} activation and batch normalization. The first two convolutional layers both have 128 output channels, the third and fourth 64 channels, each.
The convolutional layers are followed by one fully-connected layer of size 512. This model was trained using a batch size of 64.

\paragraph{Relation Network} 
\citet{santoro2017simple} introduced a simple yet powerful neural module for relational reasoning. It takes as input a set of $N$ ``object'' representations $\mathbf{o}_i\in\mathbb{R}^C, i=1,\dots,N$ and computes a representation of relations between objects according to
\begin{equation}
    \label{eq:rn}
    RN(\mathbf{O}) = f_\phi\left(\frac{1}{N^2}\sum\limits_{i,j} g_\theta(\mathbf{o}_{i,\cdot}, \mathbf{o}_{j, \cdot})\right),
\end{equation}
where $\mathbf{O}\in\mathbb{R}^{N\times C}$ is the matrix containing $N$ $C$-dimensional object representations $\mathbf{o}_{i,\cdot}$ stacked row-wise.
Both $f_\phi$ and $g_\theta$ are implemented as \glspl{mlp}, making the relational module fully-differentiable.

In our FigureQA experiments, we follow the overall architecture used by \citet{santoro2017simple} in their experiments on CLEVR from pixels, adding one convolutional layer to account for the higher resolution of our input images and increasing the number of channels. We do not use random rotations for data augmentation, to avoid distortions that might change the correct response to a question.

The object representations are provided by a \gls{cnn} with the same architecture as the one in the previous baseline, only dropping the fully-connected layer at the end. 
Each pixel of the \gls{cnn} output (64 feature maps of size $8\times8$) corresponds to one ``object'' $\mathbf{o}_{i, \cdot}\in\mathbb{R}^{64}, i\in[1,\dots,H \cdot W]$,
where $H$ and $W$, denote height and width, respectively. 
To also encode the location of objects inside the feature map, the row and column coordinates are concatenated to that representation:
\begin{equation}
    \mathbf{o}_i \leftarrow (o_{i,1}, \dots, o_{i,64}, \lfloor\frac{i-1}{W}\rfloor, (i-1) \,(\operatorname{mod} W)).
\end{equation}

The \gls{rn} takes as input the stack of all pairs of object representations, concatenated with the question; here the question encoding is once again produced by an \gls{lstm} with 256 hidden units. 
Object pairs are then separately processed by $g_\theta$ to produce a feature representation of the relation between the corresponding objects. The sum over all relational features is then processed by $f_\phi$, yielding the predicted outputs.

\begin{figure}[h]
    \centering
    \includegraphics[width=.99\textwidth]{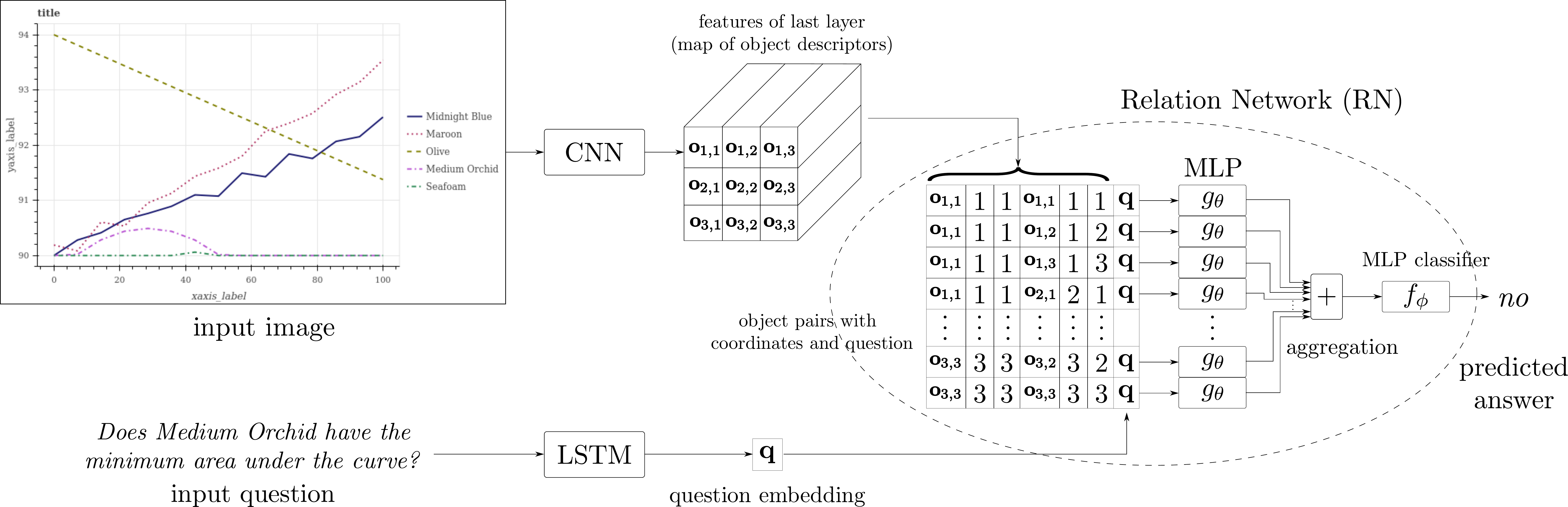}
    \caption{Sketch of the \gls{rn} baseline.}
    \label{fig:rn}
\end{figure}

The \gls{mlp} implementing $g_\theta$ has four layers, each with $256$ \gls{relu} units. The \gls{mlp} classifier $f_\phi$ processing the overall relational representation, has two hidden layers, each with $256$ \gls{relu} units, the second layer using dropout with a rate of $50\%$.
An overall sketch of the \gls{rn}'s structure is shown in Figure~\ref{fig:rn}. The model was trained using four parallel workers, each computing gradients on batches of size 160, which are then averaged for updating parameters.

\FloatBarrier

\section{Experimental Results}
\begin{figure}[h]
    \centering
    \includegraphics[width=.6\textwidth]{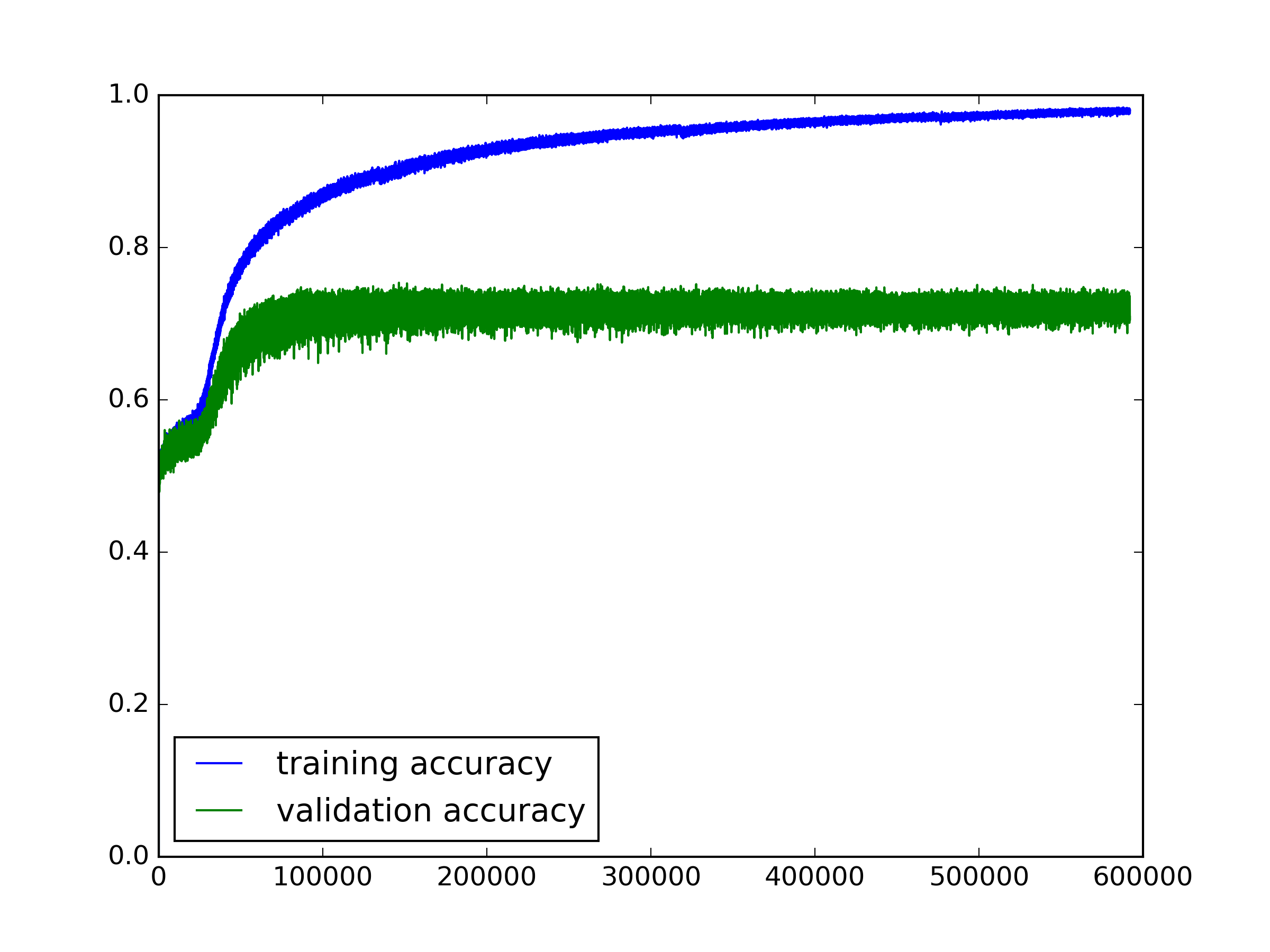}
    \caption{Learning curves of the \gls{rn}.}
    \label{fig:rn_learning_curve}
\end{figure}

All model baselines are trained and evaluated using the alternated color scheme.
At each training step, we compute the accuracy on one randomly selected batch from the validation set and keep an exponential moving average with decay $0.9$.
Starting from the 100th update, we perform early-stopping using this moving average.
The best performing model using this approximate validation performance measure is evaluated on the whole test set. Results of all our models are reported in Table~\ref{tab:acc}.
Figure~\ref{fig:rn_learning_curve} shows the training and validation accuracy over updates for the \gls{rn} model.

The comparison between text-only and CNN+LSTM models shows that the visual modality contributes to learning; however, due to the relational structure of the questions, the \gls{rn} significantly outperforms the simpler CNN+LSTM model.

Our editorial team answered a subset from our test set, containing $16,876$ questions, corresponding to $1,275$ randomly selected figures (roughly 250 per figure type).
The results are reported in Table~\ref{tab:acc_on_subset} and compared with the CNN+LSTM and \gls{rn} baselines evaluated on the same subset.
Our human baseline shows that while the problem is also challenging for humans, there is still a significant performance margin over our model baselines.

Tables~\ref{tab:fig-type} and~\ref{tab:q-type} show the performances of the CNN+LSTM and \gls{rn} baselines compared to the performances of our editorial staff by figure type and by question type, respectively.
More details on the human baseline and an analysis of results are provided in Appendix~\ref{sec:human}.

\FloatBarrier
\begin{table}[h]
    \caption{Performance of our baselines on the validation and test sets with the alternated color scheme.}
    \label{tab:acc}
    \begin{center}
        \begin{tabular}{lcc}
            \toprule
            Model & Validation Accuracy  (\%) & Test Accuracy  (\%) \\
            \midrule
            Text only & \besttextvaltwo & \besttexttesttwo \\ 
            CNN+LSTM & \bestcnnvaltwo & \bestcnntesttwo \\ 
            CNN+LSTM on VGG-16 features & \bestvggvaltwo & \bestvggtesttwo \\ 
            \gls{rn} & \bestrnvaltwo & \bestrntesttwo \\ 
            \bottomrule
        \end{tabular}
    \end{center}
\end{table}

\begin{table}[h]
    \caption{Performance of CNN+LSTM, \gls{rn} and our human annotators on a subset of the test set with the alternated color scheme.}
    \label{tab:acc_on_subset}
    \begin{center}
        \begin{tabular}{lc}
            \toprule
            Model & Test Accuracy  (\%) \\
            \midrule
            CNN+LSTM & \bestcnntesttwohuman \\
            \gls{rn} & \bestrntesttwohuman \\ 
            Human & \besthumantesttwohuman \\
            \bottomrule
        \end{tabular}
    \end{center}
\end{table}

\begin{table}[h]
    \caption{CNN+LSTM, \gls{rn} and human accuracy (in percent) per figure type on a subset of the test set with the alternated color scheme.}
    \label{tab:fig-type}
    \begin{center}
        \begin{tabular}{lccc}
            \toprule
            Figure Type & CNN+LSTM & \gls{rn} & Human \\
            \midrule
            Vertical Bar & 59.63 & 77.13 
            & 95.90 \\
            Horizontal Bar & 57.69 & 77.02 
            & 96.03 \\
            Line & 54.46 & 66.69 
            & 90.55 \\
            Dot Line & 54.19 & 69.22 
            & 87.20 \\
            Pie & 55.32 & 73.26 
            & 88.26 \\
            \bottomrule
        \end{tabular}
    \end{center}
\end{table}

\begin{table}[h]
    \caption{CNN+LSTM, \gls{rn} and human accuracy (in percent) per question type. 
    The reported accuracies are both computed on the same subset of the test set with alternated color scheme.}
    \label{tab:q-type}
    \begin{center}
        \begin{tabular}{clccc}
            \toprule
            {} & Template   & CNN+LSTM & \gls{rn} & Human \\
            \midrule
            1 & Is $X$ the minimum?         & 56.63 & 76.78 
                                                        & 97.06 \\
            2 & Is $X$ the maximum?         & 58.54 & 83.47 
                                                        & 97.18 \\
            3 & Is $X$ the low median?      & 53.66 & 66.69 
                                                        & 86.39 \\
            4 & Is $X$ the high median?     & 53.53 & 66.50 
                                                        & 86.91 \\
            5 & Is $X$ less than $Y$?       & 61.36 & 80.49 
                                                        & 96.15 \\
            6 & Is $X$ greater than $Y$?    & 61.23 & 81.00 
                                                        & 96.15 \\
            7 & Does $X$ have the minimum area under the curve? & 56.60 & 69.57 
                                                                        & 94.22 \\
            8 & Does $X$ have the maximum area under the curve? & 55.69 & 78.45 
                                                                        & 95.36 \\
            9 & Is $X$ the smoothest?       & 55.49 & 58.57 
                                                                        & 78.02 \\
            10 & Is $X$ the roughest?       & 54.52 & 56.28 
                                                                        & 79.52 \\
            11 & Does $X$ have the lowest value?     & 55.08 & 69.65 
                                                                        & 90.33 \\
            12 & Does $X$ have the highest value?    & 58.90 & 76.23 
                                                                        & 93.11 \\
            13 & Is $X$ less than $Y$?\tablefootnote{\label{note:humanstrictly} In the sense of \emph{strictly greater/less} than. This clarification is provided to judges for the human baseline.}       & 50.62 & 67.75 
                                                        & 90.12 \\
            14 & Is $X$ greater than $Y$?\footref{note:humanstrictly}      & 51.00 & 67.12 
                                                        & 89.88 \\
            15 & Does $X$ intersect $Y$?     & 49.88 & 68.75 
                                                        & 89.62 \\
            \bottomrule
        \end{tabular}
    \end{center}
\end{table}

\FloatBarrier

\section{Conclusion}
We introduced FigureQA, a machine learning corpus for the study of visual reasoning on scientific figures.
To build this dataset, we synthesized over one million question-answer pairs grounded in over $100,000$ synthetic figure images.
Questions examine plot characteristics like the extrema, area-under-the-curve, smoothness, and intersection, and require integration of information distributed spatially throughout a figure.
The corpus comes bundled with side data to facilitate the training of machine learning systems.
This includes the numerical data used to generate each figure and bounding-box annotations for all plot elements.
We studied the visual-reasoning task by training four neural baseline models on our data, analyzing their test-set performance, and comparing it with that of humans.
Results indicate that more powerful models must be developed to reach human-level performance.

In future work, we plan to test the transfer of models trained on FigureQA to question-answering on real scientific figures, and to iteratively extend the dataset either by significantly increasing the number of templates or by crowdsourcing natural-language questions-answer pairs.
We envision FigureQA as a first step to developing models that intuitively extract knowledge from the numerous figures produced by modern science.

\subsubsection*{Acknowledgments}
We thank Mahmoud Adada, Rahul Mehrotra and Marc-Alexandre Côté for technical support, as well as Adam Ferguson, Emery Fine and Craig Frayne for their help with the human baseline. This research was enabled in part by support provided by WestGrid and Compute Canada.

\bibliography{literature}
\bibliographystyle{iclr2018_conference}

\newpage
\appendix
\section{Data Samples}
\label{sec:samples}
\hyphenation{annotations}
Here we present a sample figures of each plot type (\emph{vertical bar graph}, \emph{horizontal bar graph}, \emph{line graph}, \emph{dot line graph} and \emph{pie chart}) from our dataset along with the corresponding question-answer pairs and some of the bounding boxes.

\subsection{Vertical Bar Graph}
\begin{figure}[h]
    \centering
    \caption{Vertical bar graph with question answer pairs.}
    \label{fig:vbar_qa}
    \begin{subfigure}[]{0.7\textwidth}
        \includegraphics[width=0.95\textwidth]{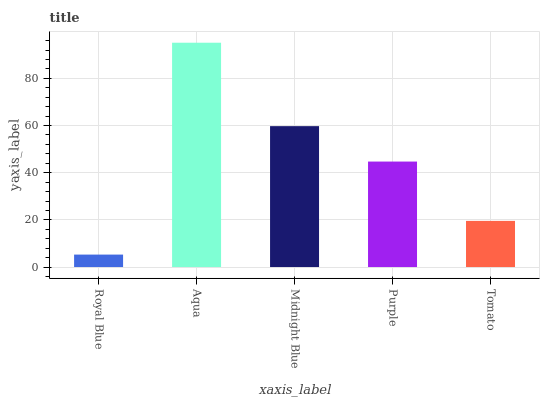}
    \end{subfigure}%
    \begin{subfigure}[]{0.3\textwidth}
        \textbf{Q:} Is Aqua the maximum?\newline
        \textbf{A: Yes}\newline\newline
        \textbf{Q:} Is Midnight Blue greater than Aqua?\newline
        \textbf{A: No}\newline\newline
        \textbf{Q:} Is Midnight Blue less than Aqua?\newline
        \textbf{A: Yes}\newline\newline
        \textbf{Q:} Is Purple the high median?\newline
        \textbf{A: Yes}\newline\newline
        \textbf{Q:} Is Tomato the low median?\newline
        \textbf{A: No}
    \end{subfigure}
\end{figure}

\begin{SCfigure}[][h]
    \includegraphics[width=0.5\textwidth]{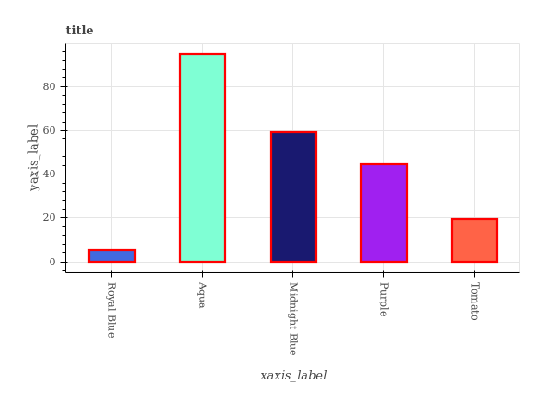}
    \caption{Vertical bar graph with some annotations.}
    \label{fig:vbar_anno}
\end{SCfigure}

\begin{SCfigure}[][h]
    \includegraphics[width=0.5\textwidth]{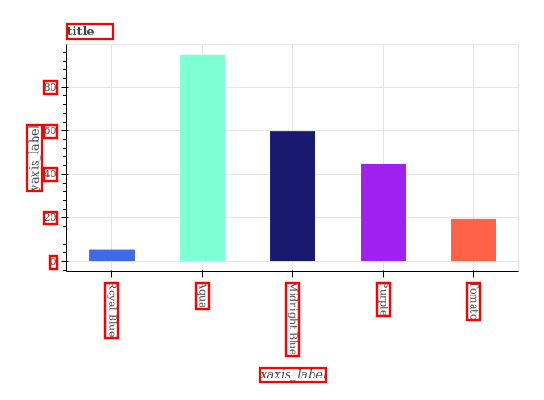}
    \caption{Vertical bar graph with label annotations.}
\end{SCfigure}

\clearpage

\subsection{Horizontal Bar Graph}
\begin{figure}[h]
    \centering
    \caption{Horizontal bar graph with question answer pairs.}
    \label{fig:hbar_qa}
    \begin{subfigure}[]{0.66\textwidth}
        \includegraphics[width=0.95\textwidth]{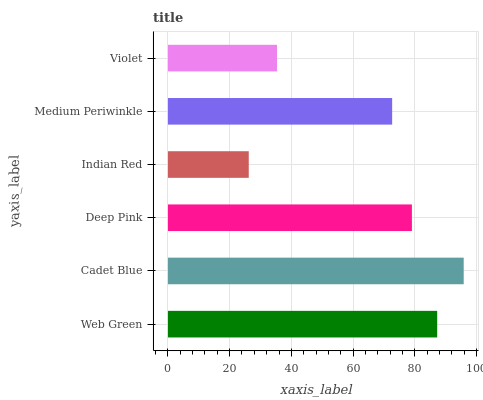}
    \end{subfigure}
    \begin{subfigure}[]{0.33\textwidth}
        \textbf{Q:} Is Deep Pink the minimum?\newline
        \textbf{A: No}\newline\newline
        \textbf{Q:} Is Cadet Blue the maximum?\newline
        \textbf{A: Yes}\newline\newline
        \textbf{Q:} Is Deep Pink greater than Cadet Blue?\newline
        \textbf{A: No}\newline\newline
        \textbf{Q:} Is Medium Periwinkle the low median?\newline
        \textbf{A: Yes}\newline\newline
        \textbf{Q:} Is Deep Pink the high median?\newline
        \textbf{A: Yes}
    \end{subfigure}
\end{figure}

\begin{SCfigure}[][h]
    \includegraphics[width=0.45\textwidth]{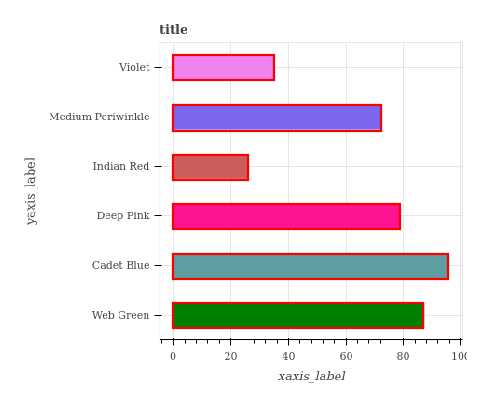}
    \caption{Horizontal bar graph with some annotations.}
    \label{fig:hbar_anno}
\end{SCfigure}

\begin{SCfigure}[][h]
    \includegraphics[width=0.45\textwidth]{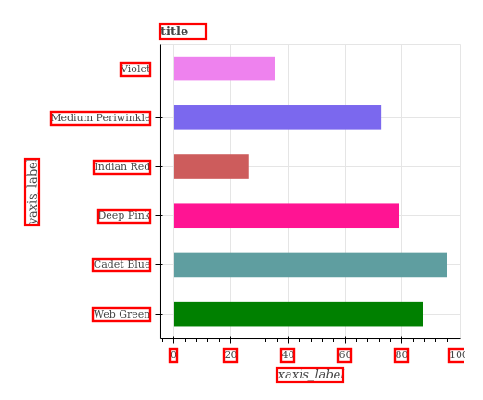}
    \caption{Horizontal bar graph with label annotations.}
    \label{fig:hbar_anno2}
\end{SCfigure}

\clearpage

\subsection{Line Graph}
\begin{figure}[h]
    \centering
    \caption{Line graph with question answer pairs.}
    \includegraphics[width=0.8\textwidth]{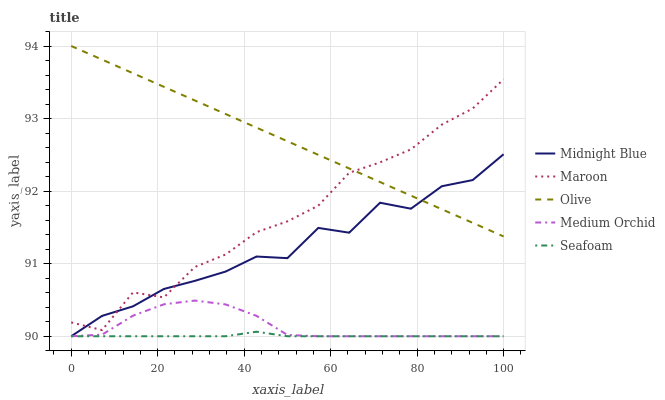}
\end{figure}

\begin{multicols}{2}
\textbf{Q:} Does Medium Orchid have the minimum area under the curve?\newline
\textbf{A: No}\newline\newline
\textbf{Q:} Is Olive the smoothest?\newline
\textbf{A: Yes}\newline\newline\newline
\textbf{Q:} Does Olive have the highest value?\newline
\textbf{A: Yes}\newline\newline
\textbf{Q:} Is Seafoam less than Olive?\newline
\textbf{A: Yes}\newline\newline
\textbf{Q:} Does Olive intersect Midnight Blue?\newline
\textbf{A: Yes}
\end{multicols}

\begin{SCfigure}[][h]
    \includegraphics[width=0.55\textwidth]{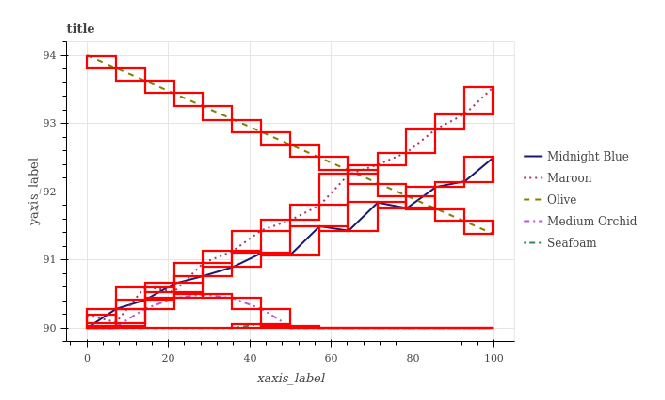}
    \caption{Line graph with some annotations.}
    \label{fig:line_anno}
\end{SCfigure}

\begin{SCfigure}[][h]
    \includegraphics[width=0.55\textwidth]{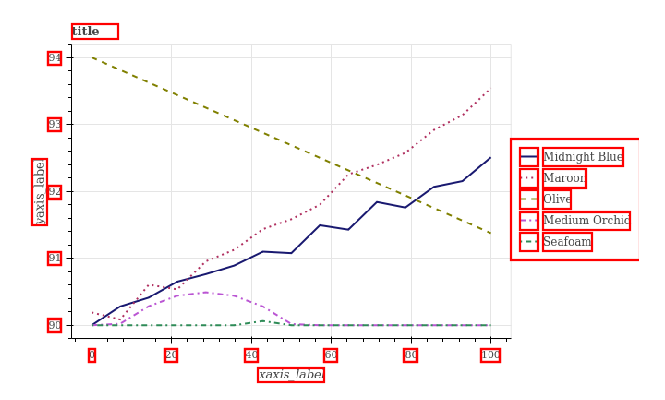}
    \caption{Line graph with label annotations.}
    \label{fig:line_anno2}
\end{SCfigure}

\clearpage

\subsection{Dot Line Graph}
\begin{figure}[h]
    \centering
    \caption{Dot line graph with question answer pairs.}
    \label{fig:dotline_qa}
    \begin{subfigure}[]{0.6\textwidth}
        \includegraphics[width=0.95\textwidth]{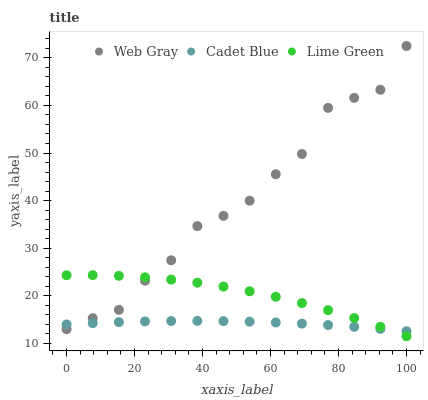}
    \end{subfigure}
    \begin{subfigure}[]{0.38\textwidth}
    \textbf{Q:} Does Web Gray have the maximum area under the curve?\newline
    \textbf{A: Yes}\newline\newline
    \textbf{Q:} Does Cadet Blue have the minimum area under the curve?\newline
    \textbf{A: Yes}\newline\newline
    \textbf{Q:} Is Web Gray the roughest?\newline
    \textbf{A: Yes}\newline\newline
    \textbf{Q:} Does Lime Green have the lowest value?\newline
    \textbf{A: Yes}\newline\newline
    \textbf{Q:} Is Lime Green less than Web Gray?\newline
    \textbf{A: No}
    \end{subfigure}
\end{figure}

\begin{SCfigure}[][h]
    \includegraphics[width=0.37\textwidth]{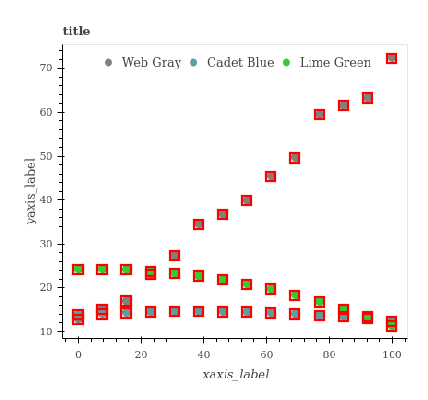}
    \caption{Dot line graph with some annotations.}
    \label{fig:dotline_anno}
\end{SCfigure}

\begin{SCfigure}[][h]
    \includegraphics[width=0.37\textwidth]{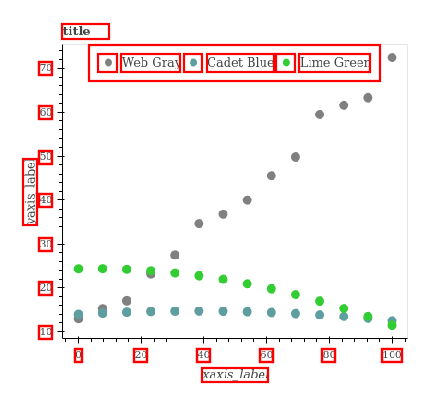}
    \caption{Dot line graph with label annotations.}
    \label{fig:dotline_anno2}
\end{SCfigure}

\clearpage

\subsection{Pie Chart}
\begin{figure}[h]
    \centering
    \caption{Pie chart with question answer pairs.}
    \label{fig:pie_qa}
    \begin{subfigure}[]{0.6\textwidth}
        \includegraphics[width=0.95\textwidth]{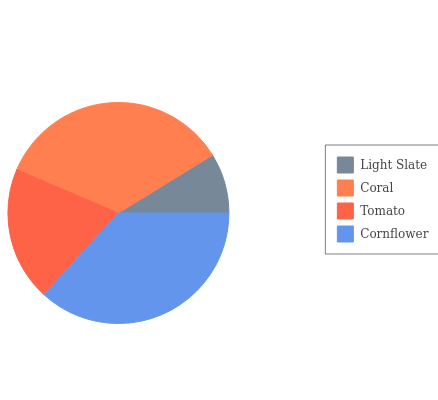}
    \end{subfigure}
    \begin{subfigure}[]{0.38\textwidth}
        \textbf{Q:} Is Coral the minimum?\newline
        \textbf{A: No}\newline\newline
        \textbf{Q:} Is Cornflower the maximum?\newline
        \textbf{A: Yes}\newline\newline
        \textbf{Q:} Is Light Slate greater than Coral?\newline
        \textbf{A: No}\newline\newline
        \textbf{Q:} Is Light Slate less than Coral?\newline
        \textbf{A: Yes}\newline\newline
        \textbf{Q:} Is Tomato the low median?\newline
        \textbf{A: Yes}
    \end{subfigure}
\end{figure}

\begin{SCfigure}[][h]
    \includegraphics[width=0.4\textwidth]{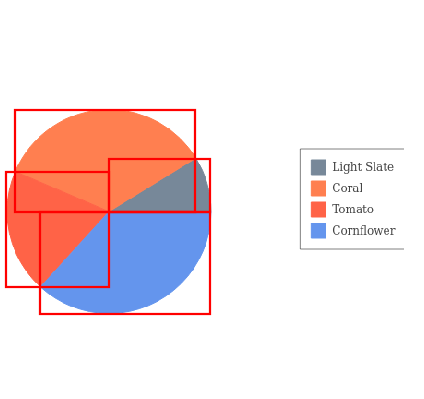}
    \caption{Pie chart with some annotations.}
    \label{fig:pie_anno}
\end{SCfigure}

\begin{SCfigure}[][h]
    \includegraphics[width=0.4\textwidth]{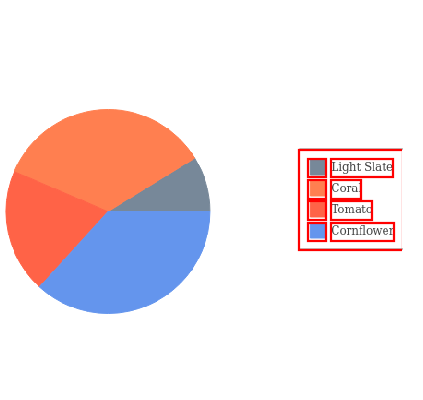}
    \caption{Pie chart with label annotations.}
    \label{fig:pie_anno2}
\end{SCfigure}

\clearpage

\section{Human baseline}
\label{sec:human}
To assess FigureQA's difficulty and to set a benchmark for model performance, we measured human accuracy on a sample of the test set with the alternated color scheme.
Our editorial staff answered $16,876$ questions corresponding to $1,275$ randomly selected figures (roughly 250 per type),
providing them in each instance with a figure image, a question, and some disambiguation guidelines.
Our editors achieved an accuracy of \besthumantesttwohuman\%, compared with \bestrntesttwohuman\% for the \gls{rn} \citep{santoro2017simple} baseline.
We provide further analysis of the human results below.

\subsection{Performance by figure type}
\begin{figure}[h]
    \centering
    \caption{Sample figures with unknown answers provided by human annotators.}
    \label{fig:unknown_answers}
    \begin{subfigure}[]{0.5\textwidth}
        \includegraphics[width=0.95\textwidth]{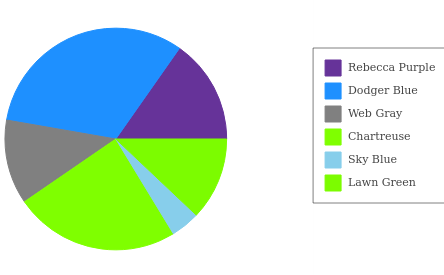}
        \textbf{Q:} Is Chartreuse the high median?\newline
    \end{subfigure}
    \begin{subfigure}[]{0.49\textwidth}
        \includegraphics[width=0.95\textwidth]{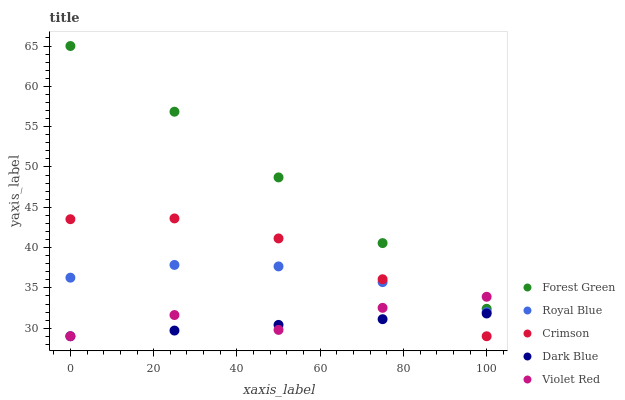}
        \textbf{Q:} Does Dark Blue intersect Royal Blue?\newline
    \end{subfigure}
\end{figure}

We stratify human accuracy by figure type in Table~\ref{tab:fig-type}.
People performed exceptionally well on bar graphs, though worse on line plots, dot-line plots, and pie charts. Analyzing the results and plot images from these figure categories, we learned that pie charts with similarly sized slices led most frequently to mistakes. Accuracy on dot-line plots was lower because plot elements sometimes obscure each other as Figure~\ref{fig:unknown_answers} shows.

\begin{figure}[h]
    \centering
    \label{fig:wrong_answers_figure}
    \caption{Sample pie chart with visually ambiguous attributes. The Sandy Brown, Web Gray, and Tan slices all have similar arc length.}
    \includegraphics[width=0.6\textwidth]{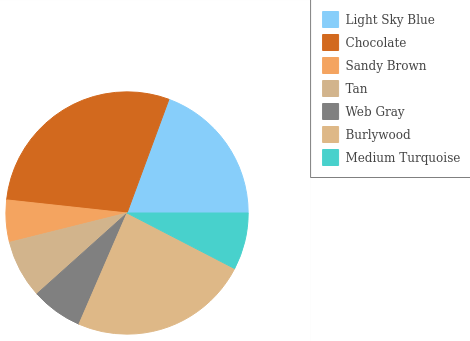}
\end{figure}

\subsection{Performance by question type}
Table~\ref{tab:q-type} shows how human accuracy varies across question types, with people performing best on minimum, maximum, and greater/less than queries.
Accuracy is generally higher on question types for categorical figures compared to continuous figures.
It is noticeably lower for questions concerning the median and curve smoothness.
Analysis indicates that many wrong answers to median questions occurred when plots had a larger number of (unordered) elements, which increases the difficulty of the task and may also induce an optical illusion.
In the case of smoothness, annotators struggled to consider both the number of deviations in a curve and the size of deviations. This was particularly evident when comparing one line with more deviations to another with larger ones.
Additionally, ground truth answers for smoothness were determined with computational or numerical precision that is beyond the capacity of human annotators. In some images, smoothness differences were too small to notice accurately with the naked eye.

\begin{figure}[h]
    \centering
    \caption{Sample figures with wrong answers illustrating common issues per question type.}
    \label{fig:wrong_answers_question}
    \begin{subfigure}[]{0.55\textwidth}
        \includegraphics[width=0.9\textwidth]{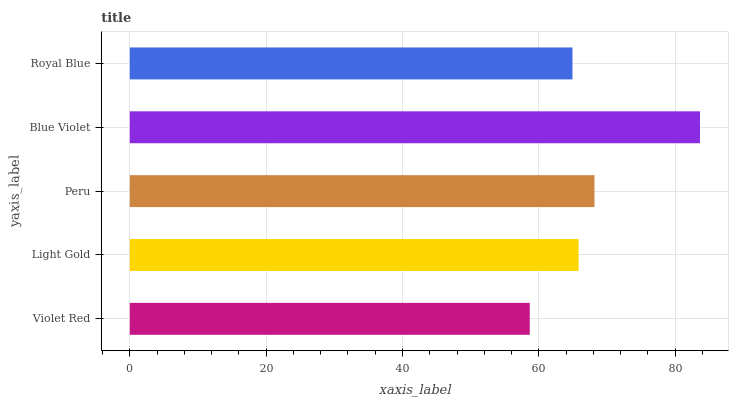}
        \newline
        Which bar is the median: Light Gold or Royal Blue?
    \end{subfigure}%
    \begin{subfigure}[]{0.45\textwidth}
        \includegraphics[width=0.9\textwidth]{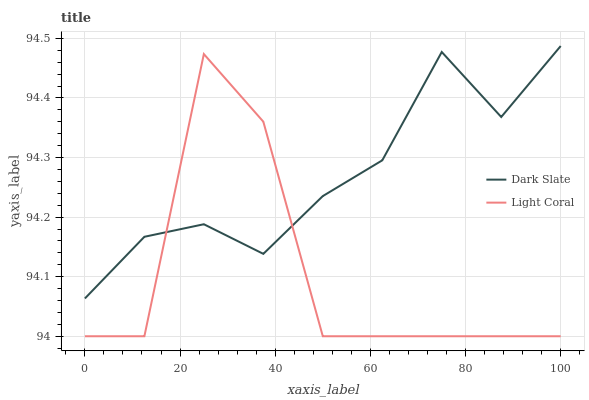}
        Which curve is rougher? One seems 'noisier' while another seems more 'jagged'.
    \end{subfigure}
\end{figure}

\FloatBarrier

\subsection{Unknown answers}
We provided our annotators with a third answer option, \emph{unknown}, for cases where it was difficult or impossible to answer a question unambiguously. Note that we instructed our annotators to select \emph{unknown} as a last resort. Only 0.34\% of test questions were answered with \emph{unknown}, and this accounted for 3.91\% of all incorrect answers. Looking at the small number of such responses, we observe that generally, annotators selected \emph{unknown} in cases where two colors were difficult to distinguish from each other, when one plot element was covered by another, or when a line plot's region of interest was obscured by a legend.

\newpage

\section{Performance of the Relation Network with and without alternated color scheme}
\label{sec:early_stop}
In this experiment we trained the \gls{rn} baseline using early stopping on both validation sets (one with the same color scheme as the training set, the other with the color-set-to-plot assignments swapped - i.e. the ``alternated'' color scheme defined in Section~\ref{sec:source_data_and_figures}), saving the respective best parameters for both. We then evaluated both models on the test sets for each color scheme. Table~\ref{tab:acc_early_stop} compares the results.
\begin{table}[h]
    \caption{Performance of our \gls{rn} baselines trained with early stopping on \emph{val1} and with early stopping on \emph{val2}. We show performances of both on \emph{test1} and \emph{test2}. The suffix \emph{``1''} denotes the training color scheme, and the suffix \emph{``2''} denotes the alternated color scheme (see Section~\ref{sec:source_data_and_figures}).}
    \label{tab:acc_early_stop}
    \begin{center}
        \begin{tabular}{lcc}
            \toprule
            Model & test1 Accuracy  (\%) & test2 Accuracy  (\%) \\
            \midrule
            \gls{rn} (val1) & \bestrnonetestone & \bestrnonetesttwo \\ 
            \gls{rn} (val2) & \bestrntestone & \bestrntesttwo \\ 
            \bottomrule
        \end{tabular}
    \end{center}
\end{table}

\end{document}